\documentclass[runningheads]{llncs}
\usepackage{graphicx}
\usepackage{xcolor}
\usepackage{amsmath}
\usepackage{subcaption}
\usepackage{multirow}
\usepackage{booktabs}

\begin{document}
\title{On Graph Neural Network Ensembles for Large-Scale Molecular Property Prediction}
\titlerunning{GNN ensemble for molecule property prediction}
\author{
Edward Elson Kosasih \inst{1, 6}\orcidID{0000-0001-5293-2641} \and
Joaquin Cabezas\inst{3, 6}\orcidID{0000-0001-7041-5082} \and
Xavier Sumba\inst{4, 6}\orcidID{0000-0002-4475-079X} \and
Piotr Bielak\inst{2,6}\orcidID{0000-0002-1487-2569} \and
Kamil Tagowski\inst{2,6}\orcidID{0000-0003-4809-3587} \and
Kelvin Idanwekhai\inst{5, 6}\orcidID{0000-0002-3804-8830} \and
Benedict Aaron Tjandra\inst{1, 6} \and 
Arian Rokkum Jamasb\inst{1, 6}\orcidID{0000-0002-6727-7579}
}
\authorrunning{Kosasih et al.}

\institute{
University of Cambridge, United Kingdom
\and
Wrocław University of Science and Technology, Wrocław, Poland \and
Universitat Rovira i Virgili, Tarragona, Spain \and
McGill University, Canada \and
Obafemi Awolowo University, Nigeria \and
ML Collective
}

\maketitle

\begin{abstract}
In order to advance large-scale graph machine learning, the Open Graph Benchmark Large Scale Challenge (OGB-LSC) \cite{hu_ogb-lsc_2021} was proposed at the KDD Cup 2021. The PCQM4M-LSC dataset  defines a molecular HOMO-LUMO property prediction task on about 3.8M graphs. In this short paper, we show our current work-in-progress solution which builds an ensemble of three graph neural networks models based on GIN, Bayesian Neural Networks and DiffPool. Our approach outperforms the provided baseline by 7.6\%. Moreover, using uncertainty in our ensemble's prediction, we can identify molecules whose HOMO-LUMO gaps are harder to predict (with Pearson's correlation of 0.5181). We anticipate that this will facilitate active learning.



\end{abstract}

\section{Introduction}
 Graph-structured data can be found in many application areas, such as social networks, molecules, protein-protein interaction networks or knowledge graphs, and are of particular interest in drug discovery \cite{gml_review}. 
 
Despite the success of graph machine learning methods, one should note that the vast majority of papers report results only for small synthetic datasets, whereas real-world datasets can contain several million or even billions of nodes \cite{hu_ogb-lsc_2021}. The most popular approach currently -- Graph Neural Networks (GNNs) -- are based on a message passing algorithm, where each node aggregates messages from its direct neighbors. However, there is a trade-off between expressivity and scalability. Therefore, applying GNNs to real-world data is a challenging task.

To address this problem, the Open Graph Benchmark Large-Scale Challenge \cite{hu2021ogb} was proposed for the KDD Cup at the KDD 2021 conference. It aims to facilitate building models that can process large-scale graphs while maintaining performance by providing three large graph datasets that encompass node, edge, and graph-level prediction tasks.

In this work, we focus on the PCQM4M-LSC quantum chemistry dataset which defines a graph regression task aiming at predicting the HOMO-LUMO gap for a given molecule. We propose an ensemble method for molecular graphs that relies on GNNs as weak learners. We demonstrate improved predictive performance that outperforms individual GNN models and other baselines by 7.6\%. We additionally perform an uncertainty analysis which shows a correlation (Pearson's R: 0.5181) between error and uncertainty.

\section{Background}

\paragraph{\textbf{HOMO-LUMO gap}} is the energy difference between the highest occupied molecular orbital (HOMO) and the lowest unoccupied molecular orbital (LUMO). Although, being able to predict  reactivity, photo-chemical or photo-physical molecule properties, the HOMO-LUMO gap cannot be determined experimentally as it is not an actual physical quantity. Traditionally this value is calculated using the Density Functional Theory (DFT) quantum mechanical calculations \cite{Huang2016}, which can take from hours to days \cite{morgante_peverati_2019}. 
This can exponentially slow down computational chemistry research on large molecule sets, limiting the ability to study applications in drug discovery, material science and photo-voltaic design.

\paragraph{\textbf{PCQM4M-LSC}} 

is a quantum chemistry dataset \cite{hu_ogb-lsc_2021}, curated under the PubChemQC project \cite{doi:10.1021/acs.jcim.7b00083}. It consists of 3,803,453 molecules and total of 55,399,880 edges. Each molecule is provided in the SMILES string format, which is further converted to a graph, along with their ground truth HOMO-LUMO gap. All molecules are identified by their PubChem ID (CID) and split into 80\% training, 10\% validation and 10\% test based on their 2D structural frameworks, ensuring that molecules in validation and test set are structurally different from training data. The regression performance is measured by the Mean Absolute Error (MAE). 

\paragraph{\textbf{OGB-LSC}} is a KDD 2021 Cup challenge. The task is to use machine learning (ML) technique on the PCQM4M-LSC dataset to predict HOMO-LUMO gaps in cheaper computational time compared to the traditional expensive DFT approach. In order for ML to be on par with DFT, it has to achieve an MAE of 0.0430.

\section{Proposed solution}
We experiment with three concepts: Bayesian networks, hierarchical clustering (graph pooling) and ensembling. 
All our solutions take into account the large scale of the dataset and the computational power required, impacting decisions regarding embedding size and depth.


\subsection{Weak learners}

\paragraph{\textbf{GIN-Virtual}} This is a baseline model that is provided as part of OGB-LSC. Graph Isomorphism Network (GIN) \cite{xu2018how} is theoretically proven to achieve maximum discriminative power among GNNs, as it generalizes the Weisfeiler-Lehman (WL) graph isomorphism test. 
We combine the GIN model with the concept of a virtual node, which is connected to every node in the graph with a special edge type. Such a setting allows information to travel long distances during the propagation phase instead of being limited to a certain number of hops within the graph. The latent dimensionality is set to 600, the number of GNN layers is set to 5. We follow the rest of the configuration provided in \cite{hu2021ogb}. The model output is clamped between 0 and 50, following physical constraints to HOMO-LUMO gaps. The L1 loss is used for training. We call this implementation \textit{GIN-Virtual}.

\begin{table}[ht]
    \centering
    \caption{MAE on validation and test splits. Only baseline and ensemble-all test values are known since only this result is released by the organisers. The MAE for each ensemble corresponds to the red bars in Figure \ref{fig:ensembling}. As seen, \textit{ensemble-all} outperforms all other models.}
    \begin{tabular}{clrrr}
        \toprule
         & Model & \#Params & Validation & Test \\
         &       &          & MAE        & MAE \\
        \midrule
        \multirow{4}{*}{Baselines \cite{hu_ogb-lsc_2021}} & GCN & 2.0M & 0.1684 & 0.1838 \\ 
        & GCN-Virtual & 4.9M & 0.1510 & 0.1579 \\ 
        & GIN & 3.8M & 0.1536 & 0.1678 \\ 
        & GIN-Virtual & 6.7M & \textbf{0.1396} & \textbf{0.1487} \\ 
        \midrule
        \multirow{4}{*}{\textbf{Ours}} & GIN-Virtual-ensemble & 6.7M & 0.1335 & n/a \\ 
        & GIN-Virtual-BNN-ensemble & 6.8M & 0.1329 & n/a \\ 
        & GIN-Virtual-DiffPool-ensemble & 7.7M & 0.1346 & n/a \\ 
        & ensemble-all & 21.2M & \textbf{0.1290} & \textbf{0.1358} \\ 
        \bottomrule
    \end{tabular}
    \label{tab:results}
\end{table}
\paragraph{\textbf{GIN-Virtual with BNN layers}} The usage of Bayesian Neural Networks (BNNs) both for message passing and the GNN readout stage improves predictive accuracy on the QM9 dataset \cite{lamb_bayesian_2020}. Particularly, Bayes by Backpropagation (BBP) \cite{blundell_weight_2015} outperforms other BNN approaches. 

In this work, we implement BBP at the readout stage using torchbnn \cite{harry_harry24kbayesian-neural-network-pytorch_2021}. We replace the  single layer neural network from the baseline GIN-Virtual with 4 Bayesian Linear layers. During training we utilize the Kullback-Leibler Divergence in addition to the L1 loss. We call this implementation \textit{GIN-Virtual-BNN}.

\paragraph{\textbf{GIN-Virtual with DiffPool}} Local pooling operators (known from Convolutional Neural Networks) are used to coarsen intermediate layer outputs. Similar approaches can be applied for GNNs, coarsening the graph into a smaller set of supernodes. Multiscale coarsening of graph data is a crucial component of scale separation priors needed as part of a geometric deep learning architecture \cite{bronstein_geometric_2021}.

In this work, we implement a recently proposed coarsening technique called DiffPool \cite{ying_hierarchical_2019}. We add a DiffPool layer after GIN message passing layers to coarsen the graph into 5 nodes. Afterwards, we run an additional GIN layer on the coarsened graph. Finally, we concatenate both graph embeddings from the original and coarsened graph, and feed them into the readout layer. We call this implementation \textit{GIN-Virtual-DiffPool}.

\subsection{GNN ensemble}
Ensembling is a technique to combine prediction outputs of several machine learning models, either with different architecture or similar, but trained from different initial conditions. In \cite{dietterich_ensemble_2000} the authors explain three possible reasons why ensembling often improves the performance compared to any single weak learner. An ensemble: (1) reduces the risk of choosing the wrong model, (2) provides better approximation as a result of training from different initial conditions and (3) forms a weighted sum representation that might not be achievable by any single weak learner.

\begin{figure}[ht]
    \centering
    \begin{subfigure}[t]{0.49\textwidth}
        \centering
        \includegraphics[width=\textwidth]{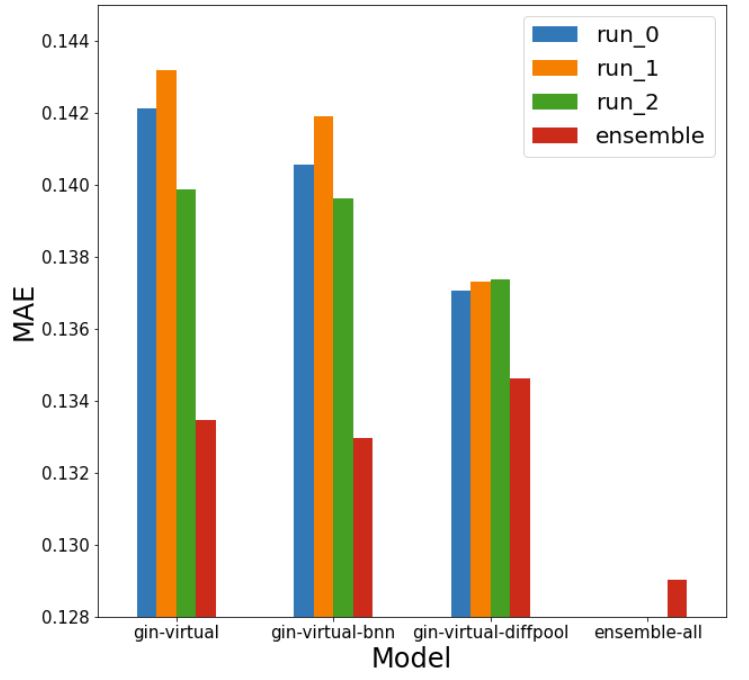}
        \caption{Ensembling significantly improves performance compared to individual weak learners. Each model is trained three times from different initial conditions (\textit{run\_\{0,1,2\}}).} \label{fig:ensembling}    
     \end{subfigure}
     \hfill
     \begin{subfigure}[t]{0.49\textwidth}
        \centering
        \includegraphics[width=\textwidth]{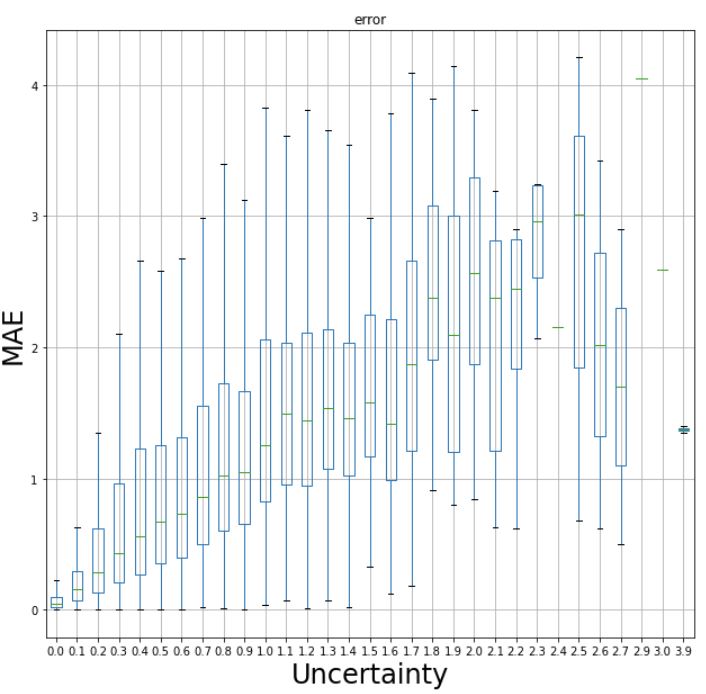}
        \caption{Prediction error tends to increase as uncertainty increases. Explain this is obtained from calculating uncertainty across the 9 different trials.} 
        \label{fig:uncertainty}
     \end{subfigure}
    \caption{Ensemble MAE results and uncertainty.}
\end{figure}

In this work, we perform two types of ensembling across the three aforementioned models. First, we train each individual model (weak learner) three times, and then take the mean of their predictions as the ensemble score. We call these, \textit{GIN-Virtual-ensemble}, \textit{GIN-Virtual-BNN-ensemble} and \textit{GIN-Virtual-DiffPool-ensemble} respectively. Second, we combine and calculate the mean of results across all models and trials. We call this \textit{ensemble-all}. 

\section{Experimental results}
Both \textit{GIN-Virtual} and \textit{GIN-Virtual-BNN} are implemented in PyTorch-Geometric \cite{fey_fast_2019}, while \textit{GIN-Virtual-DiffPool} is implemented with DGL \cite{wang_deep_2020}. We run our experiments on a Google Cloud Platform Compute Engine  machine of type n1-high-mem-16 with 16 vCPUs, 104 GB memory and 1 NVIDIA Tesla T4 GPU. Our code implementation can be found in our github\footnote{\url{https://github.com/cuent/ogb-kdd21}}.

\subsection{Baseline performance}
The PCQM4M-LSC dataset comes with a set of benchmark Graph Neural Networks models \cite{hu_ogb-lsc_2021}. Graph Convolutional Network (GCN) \cite{Kipf:2016tc} and GIN perform best in this task. Moreover, adding a virtual node \cite{pmlr-v70-gilmer17a} further improves results. Table \ref{tab:results} (upper part) shows the performance on both validation and test splits. GIN with addition of the virtual node provides the best performance.

\subsection{Ensemble performance}
All models are trained on the 3 million molecules from the training dataset and evaluated on 300,000 validation and another 300,000 test molecules. The labels in the test set are kept by the organisers of the challenge. Hence, we report full results only for the validation set. For the test set, we have only values for the baselines and our submission -- \textit{ensemble-all}. We rank 21st out of the top 49 results shown in the leaderboard\footnote{\url{https://ogb.stanford.edu/kddcup2021/results}}. 

Figure \ref{fig:ensembling} and Table \ref{tab:results} (bottom part) show that ensembling of individual types of weak learners outperforms any one of them. 
Moreover, taking an ensemble across all models and trials brought the MAE down to 0.1290. This is 7.6\% better than the reported baseline performance of 0.1396 \cite{hu_ogb-lsc_2021}.



\subsection{Uncertainty Analysis}
We also analyse whether there is any correlation between disagreement among weak learners and the task performance accuracy. We use such disagreement as a proxy for measure of uncertainty. This information, if true, would be useful to identify molecular graphs that are harder to model and predict. We anticipate that such uncertainty value could be used for future active learning tasks. 

We measure uncertainty as the standard deviation of the weak learners' predictions and find that there is indeed a tendency for performance error to increase as ensembling uncertainty goes up (see  Figure \ref{fig:uncertainty}). The Pearson's correlation of this trend is 0.5181.


\section{Conclusion}
While graph machine learning has been extensively studied recently, a vast majority of papers have only reported results for small datasets \cite{hu_ogb-lsc_2021}. To encourage development of more scalable models, the OGB-LSC challenge has been released during the KDD Cup at KDD 2021 Conference. We addressed the PCQM4M-LSC molecular property prediction task and proposed a GNN ensemble model that outperforms the provided baseline by 7.6\%. Additionally, we found that our model's uncertainty informs the error in prediction with a Pearson's correlation of 0.5181. We anticipate that this information would be useful for future active learning framework.

\section{Credit}
We would like to thank ML Collective for supporting and funding our research.

\bibliographystyle{splncs04}
\bibliography{main.bib}
\end{document}